\PassOptionsToPackage{dvipsnames}{xcolor}
\documentclass[11pt,letterpaper]{style}
\makeatletter\providecommand\@LN[2]{}\makeatother

\usepackage[numbers]{natbib}
\usepackage{graphicx}
\usepackage{booktabs}
\usepackage{multirow}
\usepackage{amsmath}
\usepackage{amssymb}
\usepackage{colortbl}
\usepackage{xcolor}
\usepackage{algorithm}
\usepackage{algpseudocode}
\usepackage{listings}
\usepackage{pgfplots}
\pgfplotsset{compat=1.17}
\usepgfplotslibrary{groupplots}
\usepackage{cleveref}
\definecolor{otblue}{HTML}{2456A6}
\definecolor{otpurple}{HTML}{8A5AA8}
\definecolor{otorange}{HTML}{C2571A}
\definecolor{otgreen}{HTML}{3D8F6B}
\definecolor{otred}{HTML}{B3261E}
\definecolor{otgray}{HTML}{63666B}
\newcommand{\mdn}{{\tiny$\downarrow$}}
\newcommand{\mup}{{\tiny$\uparrow$}}

\newcommand{\tablesize}{\small}

\newcommand{\skiptableend}{\vspace{0pt}}

\newcommand{\colfigwidth}{0.6\textwidth}  

\renewcommand\Authfont{\centering\normalfont\bfseries\fontsize{11}{15}\selectfont}
\renewcommand\Affilfont{\centering\normalfont\fontsize{10}{15}\selectfont}
\newcommand{\afflogo}[2]{%
  \raisebox{#2}{\includegraphics[height=1.5em]{#1}}%
}

\title{\centering OmniTrace: \\ A Unified Framework for Generation-Time Source Attribution in Omni-Modal LLMs}
\runningtitle{OmniTrace: A Unified Framework for Generation-Time Source Attribution in Omni-Modal LLMs}

\author{%
    {\Authfont
    \textbf{Qianqi Yan}\textsuperscript{1} \quad
    \textbf{Yichen Guo}\textsuperscript{1} \quad
    \textbf{Ching-Chen Kuo}\textsuperscript{2} \quad
    \textbf{Shan Jiang}\textsuperscript{2} \quad
    \textbf{Hang Yin}\textsuperscript{2}

    \textbf{Yang Zhao}\textsuperscript{2} \quad
    \textbf{Xin Eric Wang}\textsuperscript{1}
    }\\
    {\Affilfont
    \textsuperscript{1} \afflogo{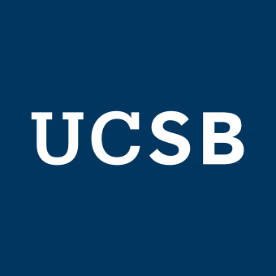}{-0.7ex} University of California, Santa Barbara \quad
    \textsuperscript{2} \afflogo{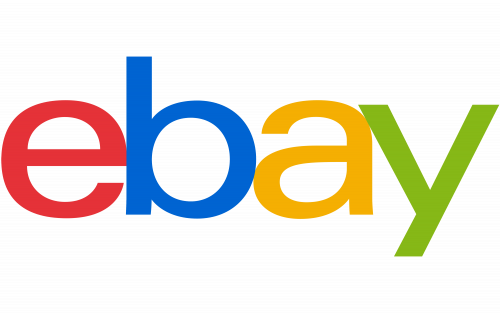}{-0.7ex} eBay \quad
    }
}

\begin{document}

\begin{abstract}
An omni-modal language model can answer a question about a multimodal document, a recording, or a video, but it cannot show which part of the input the answer came from.
Asking the model to cite its evidence yields what it \emph{says} it used; measuring what it \emph{used} has so far required rerunning it many times with parts of the input removed, a cost that grows with every second of audio and video.
We show that the generation pass already holds most of this information.
OmniTrace attributes each generated sentence to source units (text chunks, images, and seconds of audio or video) by reading late-layer attention as mass over units with no extra forward pass and no per-modality tuning; head weights can be trained where ablation data are available.
With fixed causal test setting, the units OmniTrace selects are causally important for responses and for individual sentences in video, audio, and image-text question answering on Qwen3-Omni-30B, MiniCPM-o 4.5 and Qwen2.5-Omni-7B. With one generation pass, OmniTrace recovers 77-96\% of the effect of leave-one-out attribution, which needs up to 41 extra passes.
OmniTrace also agrees best with human-marked evidence: its response-level rankings reach an AUROC of 0.83 on documents and 0.64 on video, against 0.76 and 0.60 for the strongest baseline.
\end{abstract}

\makeatletter
\g@addto@macro\theabstract{%
  \par\vskip1em\noindent{\keywordsfont Keywords: Multimodal Large Language Models, Source Attribution, Generation-Time Explanation, Model Interpretability}%
  \par\vspace{3mm}\noindent\textbf{Project Page:} \url{https://github.com/eric-ai-lab/OmniTrace}%
}
\makeatother

\maketitle

\begin{figure*}[!ht]
\setlength{\abovecaptionskip}{0.3cm}
    \centering
    \includegraphics[width=\textwidth]{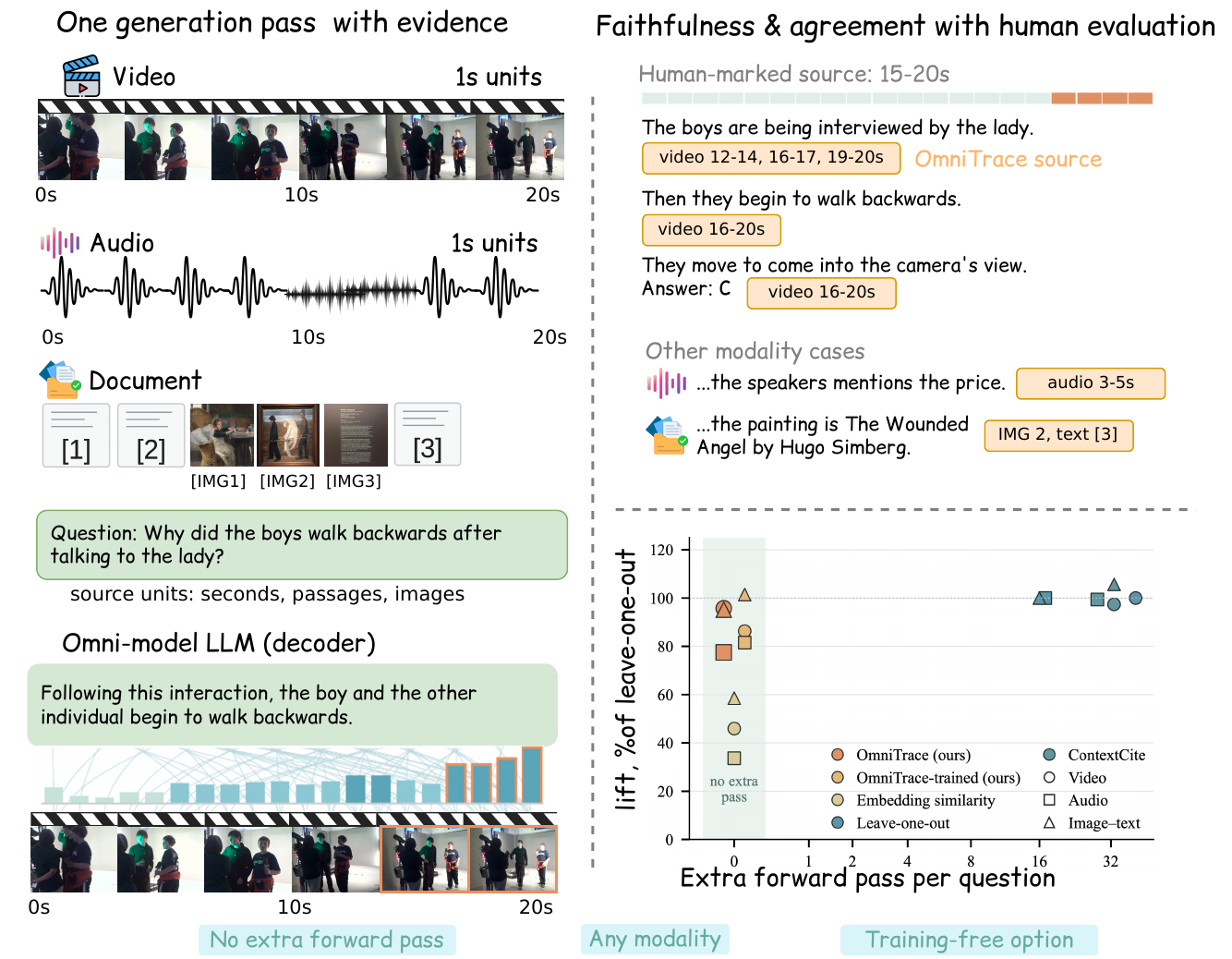}
    \caption{\textbf{OmniTrace reads what a model used from the pass that generates the answer.}
    \emph{Left:} video, audio, and documents enter one input sequence as source units (seconds, passages, images). While the decoder writes a sentence, OmniTrace sums the attention mass on each unit (bars) and labels the units that stand out within their modality (orange).
    \emph{Top right:} every sentence comes with its sources, as time intervals or source ids; here most labelled seconds fall inside the span that annotators marked, although no annotation was used.
    \emph{Bottom right:} tested by hiding units and replaying the response, OmniTrace reaches 77--96\% of the effect of leave-one-out with no extra forward pass, where perturbation needs 16 to 41 per question (Qwen2.5-Omni-7B).
    }
    \label{fig:teaser}
\end{figure*}

\section{Introduction}
\label{sec:intro}

Omni-modal large language models read text, images, audio, and video in one prompt and answer in free text~\cite{gpt6astra2026, gemini38flash2026, Xu2025Qwen3OmniTR, cui2026minicpm, xu2025qwen2, ye2026omnivinci}.
The more a model can take in, the harder its answers are to check. A claim about a forty-second clip or a pool of fifteen document quotes can only be verified by going through the input again, unless the model can show where the claim came from.

There are two ways to obtain this evidence, and they answer different questions.
One can \emph{ask} the model: prompt it to cite a time span or a passage with each sentence. This is cheap, but a citation is itself generated text. It tells us what the model says it used, and stated citations for audio and video are often missing or hallucinated~\cite{wan2026murgat}.
Or one can \emph{measure}: remove parts of the input and see how the output changes~\cite{cohen-wang2024contextcite}. This tells us what the model used, at the price of tens of extra forward passes per query, each of them over the full clip.
Learned shortcuts~\cite{cohenwang2025at2, kang2026vstream} lower the price at inference but need perturbation data for every new model, and both lines of work were developed for text contexts or single images.
To our knowledge, for audio and video there is currently no way to know what a model relied on without paying for perturbation.

Our starting point is that the model already leaves a record of where it looked.
Every generated token is computed by attending over the input, so the generation pass contains, for free, a distribution over the input for every token.
The open question is whether this record can be read faithfully once the sources are not sentences but images and seconds of sound and video.
The usual way to trace with attention follows \emph{peaks}: it links each generated token to its most attended input position.
That works when a source is a handful of text tokens. It fails for continuous sources, where one second of video is spread over hundreds of positions that each receive little weight: the second as a whole may dominate while no single position does.
Evidence in continuous modalities is diffuse, and it has to be read as \emph{mass}.

This view leads to \textbf{OmniTrace}, a training-free attribution method that runs inside the generation pass. 
As shown in \Cref{fig:teaser}, the input is split into source units by structure it already has: one-second units for audio and video and one unit per passage or image.
A unit's score for a sentence is the late-layer attention mass that the sentence's tokens place on it, and a unit is labelled as evidence when its score stands out among the units of its own modality.
The score is additive over tokens and over positions, so the same pass yields attributions for sentences and whole responses, as source ids for text and images and as time intervals for audio and video.

We replay the model's own response while source units are hidden behind content-free placeholders of the same length (black frames, silence, blank tokens), and compare the loss in probability with the loss from hiding the same number of random units or attribution produced by other baseline methods.
The same operation lets leave-one-out (Oracle) and ContextCite~\cite{cohen-wang2024contextcite} run on audio and video, which gives, to our knowledge, the first direct comparison of single-pass and perturbation-based attribution on continuous inputs.

Experiments on Qwen3-Omni-30B~\cite{Xu2025Qwen3OmniTR}, MiniCPM-o 4.5~\cite{cui2026minicpm} and Qwen2.5-Omni-7B~\cite{xu2025qwen2} on three modalities show that (1). the units OmniTrace selects are causally important for whole responses and for single sentences in all nine combinations of model and modality.
(2). On the model the setting was chosen on, one pass recovers 77-96\% of leave-one-out's effect without any of its 16 to 41 extra passes (\Cref{fig:teaser}, bottom right).
(3). What a model cites itself is not reliably what it uses: on all three models the cited parts of audio clips matter less than the parts OmniTrace finds.
(4). Training the head weights recovers 96\% of the perturbation effect, still in one pass.

Our contributions are:
\begin{itemize}
    \item \textbf{OmniTrace}, a single-pass, training-free method that attributes each generated sentence to text, image, audio, and video sources by reading attention as mass over source units, with an optional variant that trains the head weights.
    \item \textbf{A causal evaluation protocol for continuous inputs}, with content-free placeholders and size-matched controls, under which perturbation baselines apply to every modality.
    \item \textbf{Evidence on three models and three modalities}: one pass recovers 77--96\% of leave-one-out's effect, trained head weights recover 96\% of the perturbation effect where the default falls short, and OmniTrace matters more than the sources a model cites itself and agrees with human-marked evidence better than every baseline.
\end{itemize}

\section{Related Work}

\begin{table}[t]
\centering
\tablesize
\setlength{\tabcolsep}{3pt}
\begin{tabular}{lcccc}
\toprule
 & No & No extra & & Audio,\\
\textbf{Method} & training & pass & Image & video\\
\midrule
Self-citation (prompted) & \checkmark & -- & \checkmark & \checkmark\\
ContextCite~\cite{cohen-wang2024contextcite} & \checkmark & -- & -- & --\\
MIRAGE~\cite{qi-etal-2024-model} & \checkmark & -- & -- & --\\
AT2~\cite{cohenwang2025at2} & -- & \checkmark & -- & --\\
MultAttnAttrib~\cite{tran2026multattnattrib} & \checkmark & \checkmark & \checkmark & --\\
vStream~\cite{kang2026vstream} & -- & \checkmark & \checkmark & --\\
VTrace~\cite{yuan2026vtrace} & \checkmark & -- & \checkmark & --\\
\midrule
\textbf{OmniTrace} & \checkmark & \checkmark & \checkmark & \checkmark\\
\bottomrule
\end{tabular}
\caption{Attribution methods for generated text. Image and audio, video: source types the method is defined and evaluated on beyond text.}
\label{tab:position}
\skiptableend
\end{table}

\paragraph{Context attribution for text LLMs.}
A recent line of work attributes a generated statement to the context passages that support it.
ContextCite~\cite{cohen-wang2024contextcite} hides random subsets of the context and fits a sparse linear model of the response probability; ARC-JSD~\cite{li2026arcjsd} and AttriBoT~\cite{liu2025attribot} make such perturbation estimates more accurate or cheaper.
MIRAGE~\cite{qi-etal-2024-model} uses gradient saliency to link answer tokens to retrieved documents.
Closer to our signal, AT2~\cite{cohenwang2025at2} learns one weight per attention head from perturbation data, and AttnTrace~\cite{wang2026attntrace} and \citet{ding2025attention} read attention directly for text contexts.
All of these define sources as discrete text units and are evaluated on text (\Cref{tab:position}); none has been applied to audio or video.
OmniTrace uses the same kind of signal as the attention-based methods, needs no training, and is defined over units of any modality, including time intervals.

\paragraph{Attribution for vision--language models.}
vStream~\cite{kang2026vstream} streams attribution to image regions during reasoning but learns a map from attention to perturbation effects. VTrace~\cite{yuan2026vtrace} and ERCR~\cite{liang2025ercr} are post-hoc methods for image patches, and MultAttnAttrib~\cite{tran2026multattnattrib} applies selected attention heads to long documents.
These methods cover images and text. None covers audio or video timelines, and none tests attribution causally on continuous inputs.

\paragraph{Temporal evidence in audio and video.}
Models can be trained or prompted to state the time intervals that support an answer~\cite{lu2025vited, wang2025groundedvideollm, zhu2026timelens2}.
MuRGAt~\cite{wan2026murgat} shows that such stated citations are often hallucinated even when the reasoning is correct, and NExT-GQA~\cite{xiao2024nextgqa} shows that video models often answer correctly without looking at the relevant moment.
Mechanistic studies trace where audio information enters the answer across layers~\cite{cho2026tracingaudio, chen2026causalaudio}, but they do not attribute individual generated claims.
We measure which intervals a model relied on for each sentence and compare them with the intervals it states.

\paragraph{Evaluating attribution.}
Attribution can be scored by agreement with human evidence or by its causal effect on the output~\cite{deyoung2020eraser, atanasova-etal-2020-diagnostic}. The two can disagree: \citet{alhamidi2026carve} find that the sensitivity of video models to removed frames tracks annotated evidence only weakly.
We therefore use causal tests as the main evaluation and report agreement with human evidence separately.

\section{OmniTrace}
\label{sec:method}

\subsection{Problem Setting}
\label{sec:problem}

\paragraph{Input and source units.}
A decoder-only omni-modal LLM receives an input sequence $\mathbf{x}=(x_1,\dots,x_N)$ in which text tokens and the encoder outputs for images, audio, and video share one sequence.
We split the attributable part of $\mathbf{x}$ into \emph{source units} $\mathcal{S}=\{S_1,\dots,S_m\}$, each a set of input positions, using structure the input already has.
A text passage is one unit and an image is one unit.
Audio and video are cut into units of about one second, following the time grid of the model's own encoder, so each unit $S_j$ has a time interval $[b_j,e_j]$.
The system prompt, the question, the answer options, and template tokens are part of the prompt and are never candidates.
Each unit belongs to one modality, and we write $\mathcal{M}(S_j)$ for the set of units that share the modality of $S_j$.

\paragraph{Output.}
The model generates $\mathbf{y}=(y_1,\dots,y_T)$ with greedy decoding. We split $\mathbf{y}$ into sentences $C_1,\dots,C_K$, each a set of generated-token indices.
For every sentence $C_k$, and for the response as a whole, we want (i) a score for every unit that ranks the units by how much the model relied on them, and (ii) a small labelled set $\hat{\mathcal{S}}_k\subseteq\mathcal{S}$, reported as source ids for text and images and as time intervals for audio and video.
A sentence may get an empty set.

\subsection{Reading Attention as Mass over Units}
\label{sec:scoring}

\paragraph{Principle.}
When the model predicts $y_t$, each attention head distributes one unit of weight over the positions it reads from.
Let $A^{(\ell,h)}_t(i)$ be the weight on input position $i$ in head $h$ of decoder layer $\ell$.
We average over all heads and over the last third of the layers, $\mathcal{L}_{\text{late}}$:
\begin{equation}
a_t(i)=\frac{1}{|\mathcal{L}_{\text{late}}|\,H}\sum_{\ell\in\mathcal{L}_{\text{late}}}\sum_{h=1}^{H}A^{(\ell,h)}_t(i),
\end{equation}
so $a_t$ is a distribution over where the token $y_t$ reads from.
A source unit is a set of positions, and the natural question is how much of this distribution falls on the set.
OmniTrace scores a unit by this \emph{attention mass}, summed over the tokens of a sentence:
\begin{equation}
s_k(S_j)=\sum_{t\in C_k}\;\sum_{i\in S_j} a_t(i).
\label{eq:mass}
\end{equation}
The score is additive in two ways. Over tokens, the response-level score is the sum of the sentence scores, so one pass gives attributions at both levels. Over positions, the score of two merged units is the sum of their scores, so the scores do not depend on how a stream is cut.
Nothing else enters: no token weights, no normalization over sources, and no fitted parameter.

\label{sec:design}
A common way to read attention is to trace each generated token to its single highest-weight position and to count these traces as votes.
For text sources the peak and the mass tell the same story, because a source is a few tokens.
For continuous sources they do not. One second of video is spread over hundreds of positions, each of which receives little weight; the second as a whole can hold much of the mass while no position in it wins the argmax, which goes to a sharply attended text token.
We measured this on 1{,}000 questions from the dev set of NExT-GQA~\cite{xiao2024nextgqa} that share no video with our test set.
Ranking video seconds by argmax votes agrees with human-marked evidence at chance level in every layer (AUROC 0.50--0.52), where mass reaches 0.60; and when the argmax ranges over the whole prompt, the question text wins about 98\% of the tokens.

\paragraph{Cost.}
Scores are read during generation. At each step we keep only the attention row of the token being generated in the late layers, $O(|\mathcal{L}_{\text{late}}|\,N)$ memory per step.
OmniTrace adds no forward pass, no backward pass, and no training to the generation it explains.

\paragraph{A trained variant.}
The default weights the late heads equally. When ablation data for a model can be afforded, \textbf{OmniTrace-trained} replaces the average by a weighted sum over all heads, $a_t(i)=\sum_{\ell,h}w_{\ell h}A^{(\ell,h)}_t(i)$, and fits $w$ so that the scores predict the effect of hiding units (\Cref{appendix:trained}); the objective learns head weights for text contexts.
Units, the reading pass, and labels stay the same, and no pass is added at test time.

\subsection{From Scores to Evidence Labels}
\label{sec:labels}

Scores rank the units. To give the user a short answer, we label the units that stand out from their competitors.
Let $\bar{s}_k^{\,\setminus j}$ be the mean score of the other units of the same modality as $S_j$, that is, the mean of $s_k(S')$ over $S'\in\mathcal{M}(S_j)\setminus\{S_j\}$.
A unit is labelled for sentence $C_k$ when its score is at least $\tau$ times this mean:
\begin{equation}
S_j\in\hat{\mathcal{S}}_k \iff s_k(S_j)\;\ge\;\tau\,\bar{s}_k^{\,\setminus j}.
\label{eq:label}
\end{equation}
Three properties follow from this rule.
First, units compete only within their modality, so text passages cannot crowd out images, whose scores are on a different scale.
Second, the number of labelled units adapts to the sentence: flat scores give an empty label, so a sentence that draws on nothing in particular abstains.
Third, the rule is relative, so it does not depend on the absolute scale of attention, which changes with input length and model.
Adjacent labelled time units are merged into intervals $[b,e]$; text and image units are listed by id.
We use $\tau=1.5$ for every modality and every model (\Cref{appendix:threshold}).

\section{Experiments}
\label{sec:exp}

We test the attribution by intervention and organize the experiments around four questions.
\textbf{RQ1}: do the units OmniTrace selects matter to the model, in every modality?
\textbf{RQ2}: how much of the faithfulness of perturbation methods does one pass recover, and at what cost?
\textbf{RQ3}: does the model cite what it uses?
\textbf{RQ4}: does one fixed setting transfer to other models?

\subsection{A Causal Test for Every Modality}
\label{sec:protocol}

\paragraph{Replay with hidden units.}
We keep the model's own response $\mathbf{y}$ fixed and score it with teacher forcing, $\ell(\mathbf{y}\mid\mathbf{x})=\frac{1}{T}\sum_t\log P(y_t\mid\mathbf{x},y_{<t})$.
For a set of units $\mathcal{H}$, the \emph{drop} is $\Delta(\mathcal{H})=\ell(\mathbf{y}\mid\mathbf{x})-\ell(\mathbf{y}\mid\mathbf{x}_{\setminus\mathcal{H}})$ in nats per token, where $\mathbf{x}_{\setminus\mathcal{H}}$ is the input with the units in $\mathcal{H}$ hidden.

\paragraph{Hiding without a trace.}
Continuous inputs raise a problem that text does not: a second of audio cannot be deleted without shifting everything after it.
We replace a hidden unit \emph{at the input of the encoder} by a placeholder of the same length that carries no content: black frames, silence, a black image, or blank tokens.
Token counts, positions, and the time stamps of all other units stay the same.
This operation is also what lets leave-one-out and ContextCite~\cite{cohen-wang2024contextcite}, which are defined for text, run on audio and video.

\paragraph{Metrics.}
Every number is relative to hiding the same count of random units, so that the amount hidden is controlled.
\emph{Lift} is the extra drop when a method's top-ranked units are hidden, $\Delta(\text{top})-\mathbb{E}\,\Delta(\text{random})$.
\emph{Keep gain} is the reverse test, in which everything except the selected units is hidden.
At sentence level we hide the top units of one sentence and measure the drop of that sentence. \emph{Sentence lift} compares it with random units, and \emph{sentence specificity} compares it with the mean drop of the other sentences, so it is positive only if the units matter for this sentence in particular.
Budgets are 10--50\% of the duration for audio and video and 1--3 of 15 quotes for documents; tables use 20\% and 2 quotes (1 quote at sentence level).
We report 95\% bootstrap intervals and compare methods by paired differences (\Cref{appendix:causal}).

\subsection{Setup}

\begin{table}[t]
\centering
\tablesize
\setlength{\tabcolsep}{2.2pt}
\begin{tabular}{lccc}
\toprule
 & \textbf{Video} & \textbf{Audio} & \textbf{Image--text}\\
\midrule
Dataset & NExT-GQA & MMAU & MMDocRAG \\
Task & multiple choice & multiple choice & long answer \\
Questions  & 300 & 249  & 400  \\
Source unit & 1 s of frames & 1 s of audio & passage / image \\
Units per question & 46 & 22 & 15 \\
Input length & 10--90 s & 1.5--35 s & 15 quotes \\
Human evidence & time spans & -- & gold quotes \\
\midrule
\multicolumn{4}{l}{\textit{Answer accuracy (\%)}}\\
Qwen2.5-Omni-7B & 72.3 & 68.3 & 23.8$^\ast$ \\
MiniCPM-o 4.5 & 77.0 & 76.3 & 50.5$^\ast$ \\
Qwen3-Omni-30B & 78.7 & 73.1 & 54.0$^\ast$ \\
\bottomrule
\end{tabular}
\caption{Data. $^\ast$Open-ended answers graded as fully correct by an LLM judge Opus 5~\cite{anthropic2026opus5} against the reference answer (\Cref{appendix:judge}).}
\label{tab:stats}
\skiptableend
\end{table}

\begin{table*}[t]
\centering
\scriptsize
\setlength{\tabcolsep}{4.2pt}
\begin{tabular}{lrcccrcccrccc}
\toprule
 & \multicolumn{4}{c}{\textbf{Video}} & \multicolumn{4}{c}{\textbf{Audio}} & \multicolumn{4}{c}{\textbf{Image--text}} \\
\cmidrule(lr){2-5}\cmidrule(lr){6-9}\cmidrule(lr){10-13}
\textbf{Method} & Fwd & Lift & S-lift & S-spec & Fwd & Lift & S-lift & S-spec & Fwd & Lift & S-lift & S-spec \\
\midrule
Embedding similarity & 0 & 0.008 & 0.009 & 0.005 & 0 & 0.034 & 0.031 & 0.005 & 0 & 0.216 & 0.216 & 0.156 \\
Self-citation & 1 & 0.013 & 0.012 & 0.009 & 1 & 0.040 & 0.036 & 0.024 & 1 & 0.184 & 0.349 & \underline{0.317} \\
Leave-one-out$^\ast$ & 41 & 0.016 & 0.024 & 0.022 & 17 & 0.100 & \underline{0.133} & \underline{0.105} & 16 & 0.368 & 0.404 & 0.286 \\
ContextCite$^\ast$ & 33 & 0.016 & 0.027 & \underline{0.023} & 28 & 0.099 & \textbf{0.139} & \textbf{0.119} & 33 & \underline{0.389} & 0.397 & 0.285 \\
\midrule
\multicolumn{13}{>{\columncolor{gray!15}}l}{\textit{OmniTrace on Qwen2.5-Omni-7B}}\\
\textbf{OmniTrace} (ours) & 0 & 0.016 & 0.015 & 0.007 & 0 & 0.077 & 0.110 & 0.084 & 0 & 0.351 & 0.376 & 0.245 \\
\textbf{OmniTrace-trained} (ours)$^\P$ & 0 & 0.014 & 0.016 & 0.010 & 0 & 0.081 & 0.116 & 0.086 & 0 & 0.374 & 0.385 & 0.250 \\
\midrule
\multicolumn{13}{>{\columncolor{gray!15}}l}{\textit{OmniTrace on MiniCPM-o 4.5}}\\
\textbf{OmniTrace} (ours) & 0 & 0.024 & 0.027 & 0.012 & 0 & \underline{0.136} & 0.102 & 0.047 & 0 & 0.105 & 0.159 & 0.098 \\
\textbf{OmniTrace-trained} (ours)$^\P$ & 0 & \textbf{0.067} & \textbf{0.065} & \textbf{0.024} & 0 & \textbf{0.169} & 0.120 & 0.054 & 0 & \textbf{0.432} & \textbf{0.706} & \textbf{0.507} \\
\midrule
\multicolumn{13}{>{\columncolor{gray!15}}l}{\textit{OmniTrace on Qwen3-Omni-30B-A3B}}\\
\textbf{OmniTrace} (ours) & 0 & \underline{0.033} & \underline{0.032} & 0.011 & 0 & 0.112 & 0.095 & 0.065 & 0 & 0.316 & \underline{0.448} & 0.309 \\
\bottomrule
\end{tabular}

\caption{\textbf{Causal faithfulness and cost.}
Fwd: extra forward passes per question.
Lift: extra drop in the response's log-probability when a method's top units are hidden, over hiding the same number of random units; S-lift and S-spec: sentence lift and specificity.
\textbf{Bold} / \underline{underline}: largest / second largest value of a column.
$^\ast$Defined for text in the original work; run on audio, video, and images with OmniTrace's units and placeholders.
$^\P$Head weights fitted on ablation data; no pass added at test time.}
\label{tab:baselines}
\skiptableend
\end{table*}

\paragraph{Models and data.}
We use Qwen2.5-Omni-7B~\cite{xu2025qwen2}, MiniCPM-o~4.5~\cite{cui2026minicpm}, and Qwen3-Omni-30B~\cite{Xu2025Qwen3OmniTR} with greedy decoding. 

As shown in \Cref{tab:stats}, NExT-GQA~\cite{xiao2024nextgqa} provides video questions with human-marked evidence intervals; the model sees the frames only.
From MMAU~\cite{sakshi2025mmau} we take the audio questions whose answer depends on a part of the clip.
MMDocRAG~\cite{dong2025mmdocrag} gives each question 15 quotes from a long document, about ten text passages and five images, of which about 19\% are gold evidence.

\paragraph{Baselines.}
All baselines rank the same units and pass through the same test.
Two cost about as little as OmniTrace: \emph{embedding similarity} between a sentence and each unit (SigLIP, CLAP, and a text encoder), and \emph{self-citation}, a second generation that asks the model to cite time spans or quote ids.
Two are perturbation methods: \emph{leave-one-out} hides each unit in turn, and \emph{ContextCite}~\cite{cohen-wang2024contextcite} fits a sparse linear model on 32 random ablations.
Both are fitted with the hiding operation and the target of the test itself, so we treat them as a practical upper bound (\Cref{appendix:baselines}).

\subsection{RQ1: The Selected Units Matter}
\label{sec:rq1}

\begin{figure*}[t]
\centering
\includegraphics[width=\textwidth]{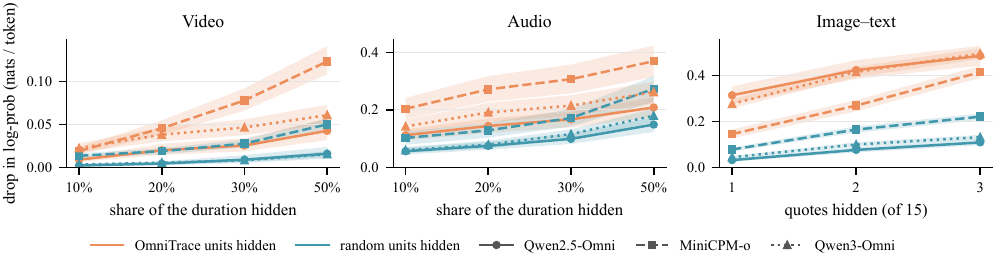}
\caption{\textbf{Hiding what OmniTrace selects hurts the response more than hiding random units of the same count}, in every modality, at every budget, and on every model. Each line is the drop in the log-probability of a model's own response when a share of its input is hidden: orange when the hidden units are the ones OmniTrace ranks highest for that model, blue when they are drawn at random. Solid lines are Qwen2.5-Omni-7B, dashed lines MiniCPM-o~4.5, and dotted lines Qwen3-Omni-30B, each explaining its own responses with the same OmniTrace setting. Bands are 95\% bootstrap intervals.}
\label{fig:curves}
\end{figure*}

\Cref{fig:curves} shows the central result.
On all three models and in all three modalities, the curve for OmniTrace's units lies above the curve for random units at every budget, and no interval of the difference includes zero (\Cref{appendix:budgets}).
The reverse test, which keeps only the selected units, agrees, reliably so except at some video budgets.

The result also holds for single sentences (\Cref{tab:baselines}, S-lift and S-spec).
Hiding the units attributed to a sentence lowers that sentence's probability more than random units do, and more than it lowers the other sentences of the response: the attribution is specific to the sentence, not a map of what is salient.
The labels of \Cref{eq:label}, which keep 10--22\% of the units, pass the deletion and sentence-level tests in all nine settings (\Cref{appendix:labels}).
Effects are large for documents, moderate for audio, and small for video, where neighboring seconds carry much of the same content.

\subsection{RQ2: Most of the Faithfulness, None of the Cost}
\label{sec:baselines}

\begin{figure}[t]
\centering
\includegraphics[width=\colfigwidth]{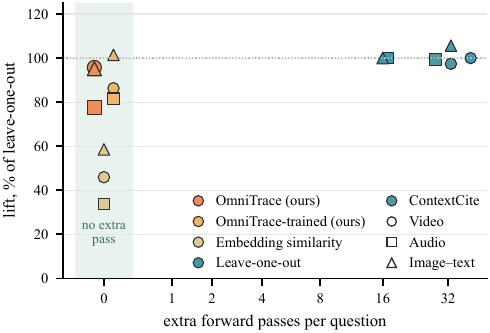}
\caption{\textbf{Faithfulness against cost}. Lift of each method as a share of leave-one-out's lift, against the extra forward passes it needs per question. Both OmniTrace variants sit in the shaded column: they read their scores from the generation pass. OmniTrace-trained needs ablation data from the same model beforehand.}
\label{fig:cost}
\end{figure}

\paragraph{Against methods of the same cost.}
OmniTrace has about twice the lift of embedding similarity in every modality (\Cref{tab:baselines}, \Cref{fig:cost}), and the gap widens at sentence level.
Units that look related to a sentence are not the units the model read: on audio its sentence specificity is close to zero.

\paragraph{Against perturbation.}
Leave-one-out and ContextCite spend 16 to 41 extra forward passes per question to measure directly what our test measures.
From a single pass, OmniTrace reaches 96\% of leave-one-out's lift on video, 77\% on audio, and 95\% on image--text input, and the share grows with the budget (\Cref{appendix:budgets}).
At sentence level it keeps 83\% and 93\% of leave-one-out's lift on audio and image--text input; video is the weak case, at about 60\%.
Perturbation remains the more faithful choice when its cost is acceptable; OmniTrace offers most of that faithfulness when it is not.

\paragraph{With trained head weights.}
OmniTrace-trained keeps the single pass and fits one weight per attention head.
On Qwen2.5-Omni the default already captures most of what training finds, 95\% and 94\% of the trained lift on audio and image--text input, and the trained weights raise image--text to 101\% of leave-one-out's lift.
On MiniCPM-o~4.5 the gain is large (\Cref{tab:baselines}): the trained lift is 0.067 on video, 0.169 on audio, and 0.432 on image--text input, which is 1.8 times leave-one-out's lift on video, 89\% of it on audio, and 96\% on image--text input, and its sentence lift on documents (0.706) is the largest in the table.
Perturbation-level faithfulness is thus available from one pass once head weights are fitted.

\paragraph{On other models.}
We repeat every baseline on MiniCPM-o~4.5 (\Cref{appendix:minicpm}).
On audio the default keeps 72--95\% of leave-one-out's lift from a 20\% budget on, and on video it overtakes leave-one-out from 30\% on.
On image--text input it reaches a quarter of the perturbation methods' lift, which the trained weights raise to 96\%; four fifths of their weight lie outside the late layers, so the generation pass of this model holds the evidence too, at other heads.
On Qwen3-Omni-30B, where every perturbation pass costs the most, the same single pass gives lifts of 0.033, 0.112, and 0.316, above Qwen2.5-Omni on video and audio.

\subsection{RQ3: Stated Evidence Is Not Used Evidence}
\label{sec:stated}

\begin{figure}[t]
\centering
\includegraphics[width=\colfigwidth]{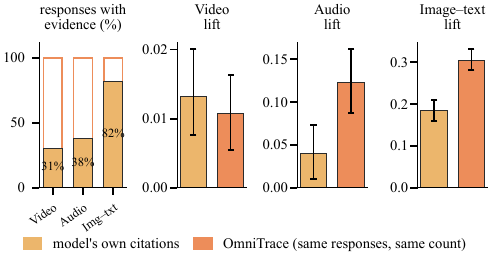}
\caption{\textbf{The model's own citations}. Left: share of responses that cite at least one valid source when asked to; OmniTrace attributes every response. Right: lift of the cited units and of OmniTrace's units of the same count on the same responses.}
\label{fig:stated}
\end{figure}

The cheapest evidence is the one the model states itself. We repeat the generation with a prompt that asks for a citation after each sentence and test the cited units (\Cref{fig:stated}).

\emph{Citations are often missing.} Qwen2.5-Omni cites a valid source in 82\% of its document responses, but in only 38\% of audio and 31\% of video responses.

\emph{Cited units matter less.} Where it does cite, hiding OmniTrace's units of the same count costs 3.1 times as much as hiding the cited units on audio and 1.7 times as much on documents; on video the two are level.
The citations also agree less with human evidence than OmniTrace's labels do (F1 of 0.22 against 0.48 on documents and 0.10 against 0.20 on video).

\emph{Citations follow the prompt.} Of the 120 audio responses that contain a time span, 32 repeat the example span of the instruction.
Asking for citations also changes the answer: video accuracy falls from 72.3\% to 55.3\%.

MiniCPM-o~4.5 and Qwen3-Omni cite more often, and their cited audio spans again matter less than OmniTrace's units (lift 0.064 against 0.098; 0.127 against 0.152), while MiniCPM-o's cited video seconds and quotes matter more (\Cref{appendix:minicpm}): how far a citation can be trusted changes with the model and the modality.
Attribution read from the generation pass leaves the response unchanged and covers every sentence.

\subsection{RQ4: One Setting, Three Models}
\label{sec:variants}

The setting was fixed on Qwen2.5-Omni video data and applied without any change to MiniCPM-o~4.5, a model from another family with a different depth and tokenization, and to Qwen3-Omni-30B, a model four times as large.
All tests of RQ1 hold on both (\Cref{fig:curves}, lower blocks of \Cref{tab:baselines}).
On Qwen3-Omni the lift on video is twice that on Qwen2.5-Omni (0.033 against 0.016), and the quotes OmniTrace selects matter more than the human-marked ones (0.316 against 0.267).

Simplicity is what makes the transfer possible.
A hard-tracing variant with per-token argmax and vote filtering is level with OmniTrace on audio and documents for the model it was developed on, but on MiniCPM-o~4.5 it loses 86\% of its lift on audio and falls behind on documents (\Cref{appendix:variants}).
Its extra steps fit one model. The plain sum of late-layer attention mass has nothing to fit.

\section{What Attention Mass Captures, and What It Misses}
\label{sec:analysis}

\begin{figure*}[t]
\centering
\includegraphics[width=\textwidth]{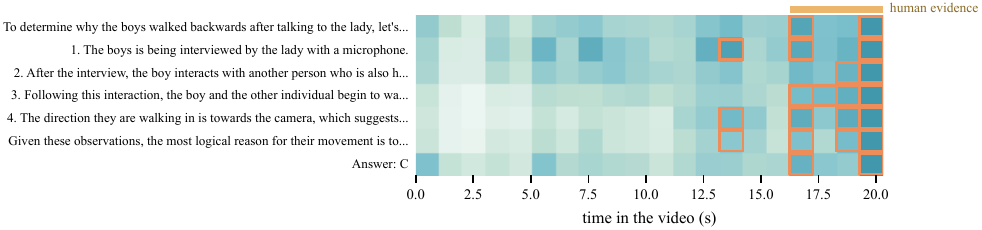}
\caption{\textbf{A response and its attribution}. Rows are generated sentences and cells are seconds of the video; darker cells hold more attention mass (scaled per row), and orange boxes are the labelled units. The bar on top is the human-marked evidence: 18 of the 21 labelled cells fall inside it, although no annotation was used.}
\label{fig:example}
\end{figure*}

\subsection{Agreement with Human Evidence}
\label{sec:agreement}

\begin{table}[t]
\centering
\tablesize
\setlength{\tabcolsep}{4pt}
\begin{tabular}{lcccc}
\toprule
 & \multicolumn{2}{c}{\textbf{Video}} & \multicolumn{2}{c}{\textbf{Image--text}}\\
\cmidrule(lr){2-3}\cmidrule(lr){4-5}
\textbf{Ranking} & AUROC & Hit@1 & AUROC & Hit@1\\
\midrule
Random & 0.500 & 0.215 & 0.500 & 0.194\\
Embedding similarity & 0.604 & 0.288 & 0.763 & 0.527\\
Leave-one-out & 0.549 & 0.309 & 0.721 & 0.657\\
ContextCite & 0.566 & \textbf{0.342} & 0.743 & 0.642\\
\textbf{OmniTrace} & \textbf{0.637} & 0.230 & \textbf{0.829} & \textbf{0.698}\\
\textbf{OmniTrace-trained} & 0.626 & 0.321 & 0.815 & 0.695\\
\bottomrule
\end{tabular}
\caption{Agreement of response-level rankings with human-marked evidence. AUROC: probability that an evidence unit is ranked above another unit. Hit@1: the top unit is evidence.}
\label{tab:agreement}
\skiptableend
\end{table}

We also compare each ranking with the human-marked evidence (\Cref{tab:agreement}; \Cref{fig:example} shows one response).

On documents, OmniTrace agrees with the gold quotes better than every baseline, including the perturbation methods. On correctly answered questions its top quote is gold in 90\% of the cases.
The same holds on Qwen3-Omni-30B: on its correctly answered document questions the AUROC is 0.91 and the top quote is gold in 90\% of the cases, and OmniTrace's labels match the gold quotes with an F1 of 0.61, against 0.39 for the model's own citations.
On video it has the best AUROC, but its single top second is evidence no more often than chance: attention finds the right region of the clip without pinning down the exact second.

The table also shows why agreement cannot be the main test: the perturbation methods have the largest causal effect and the lowest AUROC. What a model depends on and what a person marks overlap only in part.

\subsection{Unit Size and Modality Reliance}
\label{sec:size}

Units differ in size: an image quote takes hundreds of tokens and a passage a few dozen, and a larger unit can collect more mass.
On documents the scores of OmniTrace correlate with quote size at $\rho=0.51$, while the measured effects of leave-one-out and ContextCite correlate at 0.09 and 0.14 (\Cref{appendix:size}).
Dividing by the token count over-corrects ($\rho=-0.60$) and lowers the lift, so we keep the sum and report the bias.
OmniTrace-trained does not show the bias ($\rho=0.08$): a choice of heads, not a division by size, corrects it.

Across modalities the reading is accurate: images hold 74\% of the source tokens but receive 47\% of the attention mass, and leave-one-out, ContextCite, and OmniTrace-trained place 47--48\% of the measured effect on images.

The appendix adds two checks: merging one-second units into longer ones (\Cref{appendix:granularity}), and attention mass as a sign of how much a sentence depends on the sources (\Cref{appendix:flagging}).

\section{Conclusion}
What a model says it used and what it used are different things, and for audio and video only the first has been cheap to obtain.
OmniTrace narrows this gap. Reading the attention of the generation pass as mass over source units gives, at no extra cost and with one setting, attributions that hold up under intervention in three modalities on three models, and that come close to perturbation methods on the model the setting was chosen on; where it falls short, training the head weights closes most of the gap.
The causal protocol is independent of the method: with content-free placeholders and size-matched controls, any attribution method can be tested on continuous inputs.

\section*{Limitations}
\label{sec:limitations}

\paragraph{Generalizability.}
The fixed setting does not suit every model equally. On MiniCPM-o~4.5 with image-text input, OmniTrace reaches a quarter of the lift of the perturbation baselines and less than half of embedding similarity, and the human-marked quotes matter three times more than the quotes it ranks first; on video it is behind the baselines at small budgets (\Cref{appendix:minicpm}).
OmniTrace-trained closes these gaps, which shows that the evidence is in the model's attention but not in the equally weighted late layers; it needs ablation data from that model, and so gives up the training-free property.

\paragraph{Scope.}
OmniTrace needs access to attention weights and does not apply to API-only models.
The examples and tests in this paper use reasoning-style responses. Its resolution is the source unit: it finds which image and which second, not a region within an image or a word within a passage.

\bibliographystyle{unsrtnat}
\bibliography{main,related}

\newpage
\appendix

\section{Implementation Details}
\label{appendix:impl}

\paragraph{Models and decoding.}
We use the public checkpoints of Qwen2.5-Omni-7B (28 decoder layers), MiniCPM-o~4.5 (36 decoder layers), and Qwen3-Omni-30B-A3B-Instruct (48 decoder layers) in half precision (bfloat16) on two B200 GPUs, with greedy decoding, no repetition penalty, and at most 512 new tokens.
The late band of \Cref{sec:scoring} is the last third of the layers: 18--27, 24--35, and 32--47.

\paragraph{Inputs and units.}
\emph{Video.} Qwen2.5-Omni samples two frames per second within a budget of about 20k video tokens; its encoder groups two frames into one temporal step of about 1.02\,s, which is our unit. MiniCPM-o~4.5 samples one frame per second with 64 tokens per frame, and each frame is a unit. Unit times are taken from the encoder's own time grid. Videos are between 10 and 90\,s long and are given without the audio track.
\emph{Audio.} Clips are cut into one-second units along the audio encoder's token sequence.
\emph{Image--text.} Each of the 15 quotes is a unit. Quotes are given in the order of MMDocRAG, text passages first and images after them, followed by the question. Images use each model's standard image processing.
Positions outside the units (system prompt, template tokens, question, options, instruction, and whitespace) are never candidates.

\paragraph{Prompts.}
For multiple-choice questions the instruction is: ``\emph{Answer the question based on the \{video, audio\} provided. Explain your reasoning step by step. End your response with a final line of the form "Answer: X", where X is the option letter.}''
MMAU puts the correct answer first in almost half of its questions, so we shuffle the options with a fixed seed.
For the self-citation baseline the instruction adds: ``\emph{After each sentence that uses information from the \{video, audio\}, cite the time span it comes from in seconds, e.g.\ [12.0-15.5s].}'' For documents it asks for quote ids in MMDocRAG's citation format.

\paragraph{Sentences.}
Responses are split into sentences with spaCy and at line breaks. Sentences are aligned to the generated token ids, so each generated token belongs to one sentence.

\paragraph{OmniTrace-trained.}
\label{appendix:trained}
The trained variant of \Cref{sec:scoring} keeps every part of OmniTrace and changes only the head weights. The feature of a head is the attention mass it places on a unit, summed over the generated tokens of the sentence or the response, for every head of the decoder ($28\times28$ in Qwen2.5-Omni-7B), recorded in one replay of the response. With equal weights on the last third of the layers and zero elsewhere the score is the default of \Cref{eq:mass}, and training starts from these weights.
The training signal is 32 random ablations per question, in which each unit is hidden with probability 0.5 by the placeholders of \Cref{sec:protocol}. The loss is one minus the Pearson correlation between the predicted and the measured log-probability over the ablations, for responses and their sentences, minimized with Adam and early stopping on 15\% of the training questions. This objective follows AT2~\cite{cohenwang2025at2}, which is defined for text sources and has no released weights for omni-modal models; the features, the units, the ablations, and all weights here are ours.
We cross-fit in five folds, so each question is scored with weights trained on the other four.
Training costs 32 ablation passes for every training question, once per model; nothing is added at test time.
On held-out questions of Qwen2.5-Omni the trained score predicts the ablation outcomes about as well as the default on video (correlation 0.33 against 0.34) and better on audio (0.61 against 0.58) and image--text input (0.85 against 0.78). About a third of the absolute weight stays on the late layers, and about half of the weights are negative.
For labels we clip negative scores at zero and apply \Cref{eq:label} unchanged. These labels agree with human evidence better than the default's (F1 of 0.30 against 0.20 on video and 0.50 against 0.48 on documents for Qwen2.5-Omni; 0.30 against 0.14 and 0.56 against 0.26 for MiniCPM-o~4.5); we did not test them causally.

\paragraph{Reading attention.}
We run the model with its standard scaled-dot-product attention and recompute, in the late layers only, the attention row of the last position at each step. Against the attention maps returned by the framework, the largest absolute difference is $2\times10^{-7}$, and the generated tokens are identical.

\paragraph{Development study.}
The observations in \Cref{sec:design} come from 1{,}000 NExT-GQA questions on 637 videos. The 300 test questions are drawn from other videos.
Agreement with human spans was measured on the answer sentences of the 655 correctly answered questions.
We then compared the candidate settings with the causal test on 200 of these questions and fixed the setting of \Cref{eq:mass} before running any test question, any audio or image--text question, and the second model.

\section{Details of the Causal Test}
\label{appendix:causal}

\paragraph{Placeholders.}
A hidden video unit is replaced by black frames, an audio unit by silence at the clip's noise floor, an image by a black image of the same size, and a text passage by blank tokens of the same count. We verified on video that only the tokens of the hidden unit change.

\paragraph{Budgets and controls.}
For a budget $b$ on audio and video we hide the top-ranked units up to a share $b$ of the duration; for documents we hide the top $k$ quotes. Random controls hide the same number of units, with three draws per question. Keep tests hide all candidate units except the selected ones, with the same controls.
For the sentence-level test we take the top units by the sentence's own scores.
Intervals are 95\% percentile intervals from 1{,}000 bootstrap resamples of the questions (of the sentences for sentence-level metrics).

\paragraph{Two evaluation runs.}
The baselines were evaluated in a second run that tests OmniTrace again next to them. The drop for OmniTrace's top units is identical in the two runs; the random draws differ. This is why the lift of OmniTrace differs slightly between \Cref{tab:baselines} and \Cref{tab:models} (for example 0.077 and 0.069 on audio). Each paired comparison uses rows from one run.

\paragraph{Coverage.}
A few questions ran out of GPU memory in the replay and are left out: on MiniCPM-o~4.5, 13 of 246 video and 9 of 196 audio questions; on Qwen2.5-Omni, 2 video questions in the label test and 16 in the self-citation test.

\subsection{Masking Choice}
\label{appendix:masking}

\begin{table*}[t]
\centering
\small
\begin{tabular}{llcccc}
\toprule
\textbf{Model} & \textbf{Modality} ($n$) & Budget & Lift, empty & Lift, mean & Mean $-$ empty (paired) \\
\midrule
Qwen2.5-Omni-7B & Video (246) & 10\% & 0.007 & 0.007 & 0.000 {\tiny[$-$0.002, 0.003]} \\
 &  & 20\% & 0.016 & 0.014 & $-$0.002 {\tiny[$-$0.004, 0.000]} \\
 &  & 30\% & 0.017 & 0.017 & 0.000 {\tiny[$-$0.001, 0.001]} \\
 &  & 50\% & 0.027 & 0.025 & $-$0.002 {\tiny[$-$0.008, 0.003]} \\
 & Audio (196) & 10\% & 0.056 & 0.062 & 0.006 {\tiny[$-$0.002, 0.013]} \\
 &  & 20\% & 0.069 & 0.078 & 0.009 {\tiny[0.000, 0.018]} \\
 &  & 30\% & 0.068 & 0.076 & 0.008 {\tiny[0.001, 0.016]} \\
 &  & 50\% & 0.060 & 0.072 & 0.012 {\tiny[$-$0.010, 0.033]} \\
\midrule
MiniCPM-o 4.5 & Video (233) & 10\% & 0.006 & 0.007 & 0.002 {\tiny[$-$0.001, 0.004]} \\
 &  & 20\% & 0.027 & 0.028 & 0.001 {\tiny[$-$0.001, 0.004]} \\
 &  & 30\% & 0.050 & 0.054 & 0.004 {\tiny[0.001, 0.007]} \\
 &  & 50\% & 0.073 & 0.076 & 0.003 {\tiny[$-$0.001, 0.007]} \\
 & Audio (187) & 10\% & 0.102 & 0.098 & $-$0.004 {\tiny[$-$0.020, 0.011]} \\
 &  & 20\% & 0.143 & 0.138 & $-$0.005 {\tiny[$-$0.018, 0.009]} \\
 &  & 30\% & 0.135 & 0.136 & 0.001 {\tiny[$-$0.014, 0.016]} \\
 &  & 50\% & 0.097 & 0.110 & 0.014 {\tiny[0.003, 0.025]} \\
\bottomrule
\end{tabular}

\caption{Lift of OmniTrace under two ways of hiding a unit, on the same questions: a content-free placeholder at the input (\emph{empty}, used throughout the paper) and replacing the unit's embeddings by their mean (\emph{mean}).}
\label{tab:masking}
\end{table*}

\Cref{tab:masking} repeats the response-level test with the mean-embedding replacement.
On video and audio the lifts under the two fills are close for both models, most paired differences include zero, and the others are small.
We do not use the mean replacement for image--text input. A text passage whose tokens are all replaced by the same mean embedding is far from anything the model has seen: hiding one random passage this way costs 1.70 nats per token regardless of the passage, against 0.03 with blank tokens. Such a test measures the damage of the replacement, not the role of the passage.

\subsection{Replay Agreement}
\label{appendix:replay}

With greedy decoding, the token with the highest probability in the replay should be the generated token.
For MiniCPM-o~4.5 this holds at 97--98\% of the positions. For Qwen2.5-Omni-7B it holds at 88\% (video), 86\% (audio), and 81\% (image--text).
The gap comes from the generation path in half precision: on a sample of audio questions, a response generated in single precision is replayed with full agreement, and replaying a half-precision response gives the same values in half and in single precision, so the replay itself is accurate.
All drops in the paper compare two replays of the same tokens, with and without hidden units, so they are internally consistent.

\section{Baselines}
\label{appendix:baselines}

All baselines write one score per unit for the response and for each sentence, and are then evaluated by the same code as OmniTrace.

\paragraph{Embedding similarity.}
A unit is scored by the cosine similarity between the sentence and the unit. Video frames and images are embedded with SigLIP (\texttt{siglip-so400m-patch14-384}), one-second audio slices with CLAP (\texttt{clap-htsat-unfused}), and text passages with a text encoder (\texttt{bge-base-en-v1.5}). A video unit is the mean over its frames. Image and text similarities are on different scales, so for documents they are standardized within each modality. The response-level score is the mean over sentences.

\paragraph{Self-citation.}
The model is run again with the citation prompt of \Cref{appendix:impl}. A cited time span selects the units it overlaps, and a cited quote id selects that quote. Because the responses of this run differ from the main run, we compare the cited units with OmniTrace's top units of the same count on the responses of the citation run. Responses without a valid citation have no cited set and are not scored.

\paragraph{Leave-one-out.}
Each unit is hidden alone, and its score is the drop in the mean log-probability of the response or the sentence. The cost is one forward pass per unit (41, 17, and 16 on average for video, audio, and image--text).

\paragraph{ContextCite.}
We draw 32 random ablations in which each unit is kept with probability 0.5, and fit a Lasso model (regularization 0.01) that predicts the log-probability of the response or sentence from the kept units; its weights are the scores.

Both perturbation baselines use the placeholders and the teacher-forced target of the causal test itself.

\paragraph{Modalities.}
Leave-one-out is a generic procedure and ContextCite is defined for text contexts, where a source is removed by deleting its tokens. Neither has a published form for audio, video, or mixed image--text input. In this paper both run on the source units of OmniTrace and remove a unit with the content-free placeholders of \Cref{sec:protocol}; their numbers on audio, video, and images exist only through this extension.

\section{Additional Results}
\label{appendix:additional}

\subsection{Baselines on MiniCPM-o 4.5}
\label{appendix:minicpm}

\begin{table*}[t]
\centering
\small
\setlength{\tabcolsep}{4.2pt}
\resizebox{\textwidth}{!}{
\begin{tabular}{lrcccrcccrccc}
\toprule
 & \multicolumn{4}{c}{\textbf{Video} ($n$ = 246)} & \multicolumn{4}{c}{\textbf{Audio} ($n$ = 196)} & \multicolumn{4}{c}{\textbf{Image--text} ($n$ = 400)} \\
\cmidrule(lr){2-5}\cmidrule(lr){6-9}\cmidrule(lr){10-13}
\textbf{Method} & Fwd & Lift & S-lift & S-spec & Fwd & Lift & S-lift & S-spec & Fwd & Lift & S-lift & S-spec \\
\midrule
\multicolumn{13}{>{\columncolor{gray!15}}l}{\textit{Baselines on MiniCPM-o 4.5}}\\
Embedding similarity & 0 & 0.030 & 0.032 & 0.013 & 0 & 0.067\mdn & 0.055\mdn & 0.015\mdn & 0 & 0.248\mup & 0.365\mup & 0.282\mup \\
Self-citation$^\dagger$ & 1$^\ddagger$ & 0.026\mup & 0.037\mup & \textbf{0.034}\mup & 1$^\ddagger$ & 0.064\mdn & 0.093\mdn & \textbf{0.114}\mdn & 1$^\ddagger$ & 0.267\mup & \textbf{0.907}\mup & \textbf{0.903}\mup \\
Leave-one-out$^\ast$ & 41 & 0.037\mup & 0.039\mup & 0.027\mup & 17 & \underline{0.189}\mup & \underline{0.152}\mup & 0.086\mup & 16 & \underline{0.451}\mup & \underline{0.736}\mup & \underline{0.545}\mup \\
ContextCite$^\ast$ & 33 & \underline{0.050}\mup & \underline{0.058}\mup & \underline{0.032}\mup & 28 & \textbf{0.193}\mup & \textbf{0.157}\mup & \underline{0.101}\mup & 33 & \textbf{0.453}\mup & 0.722\mup & 0.542\mup \\
\midrule
\multicolumn{13}{>{\columncolor{gray!15}}l}{\textit{OmniTrace on MiniCPM-o 4.5}}\\
\textbf{OmniTrace} (ours) & 0 & 0.024 & 0.027 & 0.012 & 0 & 0.136 & 0.102 & 0.047 & 0 & 0.105 & 0.159 & 0.098 \\
\textbf{OmniTrace-trained} (ours)$^\P$ & 0 & \textbf{0.067}\mup & \textbf{0.065}\mup & 0.024\mup & 0 & 0.169\mup & 0.120\mup & 0.054 & 0 & 0.432\mup & 0.706\mup & 0.507\mup \\
\bottomrule
\end{tabular}
}
\caption{The comparison of \Cref{tab:baselines} repeated on MiniCPM-o~4.5: every baseline and both OmniTrace variants are run on this model. Columns as in \Cref{tab:baselines}. \mdn / \mup: reliably lower / higher than OmniTrace on this model on the same questions (paired 95\% bootstrap interval). $^\dagger$Self-citation is a separate generation ($^\ddagger$one extra pass), scored on the 210, 92, and 227 responses that cite; the document citation run covers 252 of the 400 questions.}
\label{tab:baselines_minicpm}
\end{table*}

\Cref{tab:baselines_minicpm} repeats the baseline comparison on the second model, with the setting of OmniTrace unchanged.

\emph{Audio.} OmniTrace has 1.4 to 2 times the lift of embedding similarity at every budget, and reaches 52\%, 72\%, 75\%, and 95\% of leave-one-out's lift at the four budgets and 62--92\% of OmniTrace-trained's.
Where the model cites, OmniTrace's units of the same count matter 1.5 times as much as the cited spans at response level and twice as much at sentence level.

\emph{Video.} The result depends on the budget. As a share of leave-one-out's lift OmniTrace reaches 27\%, 65\%, 147\%, and 194\% at the four budgets, and from 30\% on it is level with ContextCite. Its keep gain is reliably above that of embedding similarity from 20\% on. OmniTrace-trained is ahead of every method at every budget.

\emph{Image--text.} OmniTrace reaches 23\% of the lift of leave-one-out and ContextCite at two quotes (39\% at three) and 42\% of embedding similarity. This is the setting in which the fixed reading transfers worst.

\emph{What training finds.} OmniTrace-trained reads the same attention with fitted head weights. It stays within 11\% of leave-one-out on audio and documents and is the best method on video, at no test-time cost. On held-out questions its score predicts the ablation outcomes far better than the fixed setting (correlation 0.45, 0.58, and 0.94 against 0.23, 0.24, and 0.52 for video, audio, and documents), and only 18--21\% of its absolute weight lies on the late layers, against 35\% on Qwen2.5-Omni.
The evidence is therefore present in the attention of this model, in other layers than the ones the fixed setting sums. A rule that selects layers or heads without perturbation data would remove this gap; we leave it open.

\emph{Citations.} MiniCPM-o~4.5 cites in 85\%, 51\%, and 90\% of its video, audio, and document responses. Its cited video seconds and quotes have a larger lift than OmniTrace's units of the same count, and its cited audio spans a smaller one. With the citation prompt its accuracy changes from 77.0\% to 75.7\% on video and from 76.3\% to 70.7\% on audio.

\subsection{All Budgets}
\label{appendix:budgets}

\begin{table*}[t]
\centering
\small
\resizebox{\textwidth}{!}{
\begin{tabular}{llcccccc}
\toprule
\textbf{Model} & \textbf{Modality} ($n$) & Budget & Drop, top & Drop, random & Lift & Lift $>0$ & Keep gain \\
\midrule
Qwen2.5-Omni-7B & Video (246) & 10\% & 0.009 & 0.002 & 0.007 {\tiny[0.005, 0.009]} & 70\% & 0.022 {\tiny[0.006, 0.048]} \\
 &  & 20\% & 0.020 & 0.004 & 0.016 {\tiny[0.011, 0.021]} & 73\% & 0.010 {\tiny[$-$0.002, 0.021]} \\
 &  & 30\% & 0.025 & 0.009 & 0.017 {\tiny[0.013, 0.021]} & 77\% & 0.006 {\tiny[$-$0.004, 0.013]} \\
 &  & 50\% & 0.043 & 0.016 & 0.027 {\tiny[0.021, 0.034]} & 88\% & 0.005 {\tiny[$-$0.000, 0.011]} \\
 & Audio (196) & 10\% & 0.112 & 0.056 & 0.056 {\tiny[0.043, 0.069]} & 72\% & 0.042 {\tiny[0.033, 0.053]} \\
 &  & 20\% & 0.144 & 0.075 & 0.069 {\tiny[0.055, 0.086]} & 75\% & 0.066 {\tiny[0.051, 0.080]} \\
 &  & 30\% & 0.167 & 0.099 & 0.068 {\tiny[0.054, 0.083]} & 77\% & 0.073 {\tiny[0.058, 0.090]} \\
 &  & 50\% & 0.208 & 0.149 & 0.060 {\tiny[0.046, 0.073]} & 73\% & 0.080 {\tiny[0.063, 0.098]} \\
 & Image--text (400) & 1 quote & 0.314 & 0.031 & 0.282 {\tiny[0.250, 0.319]} & 92\% & 0.356 {\tiny[0.309, 0.407]} \\
 &  & 2 quotes & 0.423 & 0.076 & 0.348 {\tiny[0.310, 0.383]} & 95\% & 0.412 {\tiny[0.371, 0.456]} \\
 &  & 3 quotes & 0.485 & 0.109 & 0.377 {\tiny[0.338, 0.418]} & 94\% & 0.396 {\tiny[0.358, 0.438]} \\
\midrule
MiniCPM-o 4.5 & Video (233) & 10\% & 0.019 & 0.013 & 0.006 {\tiny[0.002, 0.010]} & 55\% & 0.010 {\tiny[$-$0.000, 0.020]} \\
 &  & 20\% & 0.046 & 0.019 & 0.027 {\tiny[0.019, 0.034]} & 73\% & 0.026 {\tiny[0.018, 0.034]} \\
 &  & 30\% & 0.078 & 0.028 & 0.050 {\tiny[0.039, 0.062]} & 79\% & 0.031 {\tiny[0.024, 0.039]} \\
 &  & 50\% & 0.123 & 0.050 & 0.073 {\tiny[0.060, 0.087]} & 85\% & 0.021 {\tiny[0.016, 0.026]} \\
 & Audio (187) & 10\% & 0.204 & 0.102 & 0.102 {\tiny[0.077, 0.131]} & 63\% & 0.060 {\tiny[0.039, 0.081]} \\
 &  & 20\% & 0.272 & 0.129 & 0.143 {\tiny[0.113, 0.174]} & 72\% & 0.106 {\tiny[0.079, 0.133]} \\
 &  & 30\% & 0.307 & 0.172 & 0.135 {\tiny[0.107, 0.162]} & 68\% & 0.128 {\tiny[0.101, 0.158]} \\
 &  & 50\% & 0.370 & 0.273 & 0.097 {\tiny[0.074, 0.121]} & 65\% & 0.125 {\tiny[0.098, 0.151]} \\
 & Image--text (400) & 1 quote & 0.145 & 0.078 & 0.067 {\tiny[0.048, 0.089]} & 58\% & 0.039 {\tiny[0.023, 0.057]} \\
 &  & 2 quotes & 0.269 & 0.164 & 0.105 {\tiny[0.076, 0.130]} & 64\% & 0.074 {\tiny[0.052, 0.096]} \\
 &  & 3 quotes & 0.414 & 0.221 & 0.194 {\tiny[0.167, 0.222]} & 75\% & 0.154 {\tiny[0.131, 0.178]} \\
\midrule
Qwen3-Omni-30B-A3B & Video (246) & 10\% & 0.022 & 0.003 & 0.019 {\tiny[0.014, 0.024]} & 70\% & 0.019 {\tiny[0.012, 0.027]} \\
 &  & 20\% & 0.038 & 0.005 & 0.033 {\tiny[0.025, 0.040]} & 74\% & 0.016 {\tiny[0.011, 0.021]} \\
 &  & 30\% & 0.047 & 0.008 & 0.038 {\tiny[0.030, 0.047]} & 76\% & 0.010 {\tiny[0.006, 0.015]} \\
 &  & 50\% & 0.061 & 0.015 & 0.046 {\tiny[0.037, 0.055]} & 77\% & 0.006 {\tiny[0.004, 0.009]} \\
 & Audio (196) & 10\% & 0.142 & 0.060 & 0.083 {\tiny[0.065, 0.103]} & 75\% & 0.059 {\tiny[0.044, 0.075]} \\
 &  & 20\% & 0.191 & 0.079 & 0.112 {\tiny[0.088, 0.137]} & 78\% & 0.097 {\tiny[0.074, 0.118]} \\
 &  & 30\% & 0.215 & 0.116 & 0.099 {\tiny[0.078, 0.125]} & 76\% & 0.100 {\tiny[0.078, 0.126]} \\
 &  & 50\% & 0.262 & 0.180 & 0.082 {\tiny[0.065, 0.100]} & 75\% & 0.094 {\tiny[0.076, 0.114]} \\
 & Image--text (400) & 1 quote & 0.276 & 0.044 & 0.232 {\tiny[0.213, 0.252]} & 96\% & 0.209 {\tiny[0.188, 0.229]} \\
 &  & 2 quotes & 0.415 & 0.099 & 0.316 {\tiny[0.294, 0.338]} & 98\% & 0.304 {\tiny[0.284, 0.329]} \\
 &  & 3 quotes & 0.495 & 0.131 & 0.363 {\tiny[0.342, 0.386]} & 99\% & 0.355 {\tiny[0.332, 0.379]} \\
\bottomrule
\end{tabular}
}
\caption{Response-level results of OmniTrace at every budget. Drops are in nats per token. Lift $>0$: share of the questions with a positive lift. Brackets: 95\% bootstrap intervals.}
\label{tab:budget}
\end{table*}

\begin{table*}[t]
\centering
\small
\setlength{\tabcolsep}{4pt}
\resizebox{\textwidth}{!}{
\begin{tabular}{llccccccccc}
\toprule
 & & & \multicolumn{4}{c}{Lift} & \multicolumn{4}{c}{OmniTrace $-$ method (paired)} \\
\cmidrule(lr){4-7}\cmidrule(lr){8-11}
\textbf{Modality} & Budget & OmniTrace & Embedding & Leave-one-out & ContextCite & OT-trained & Embedding & Leave-one-out & ContextCite & OT-trained \\
\midrule
Video & 10\% & 0.007 & 0.002 & 0.014 & 0.013 & 0.006 & 0.005 {\tiny[0.002, 0.008]} & $-$0.007 {\tiny[$-$0.011, $-$0.005]} & $-$0.006 {\tiny[$-$0.008, $-$0.004]} & 0.001 {\tiny[$-$0.001, 0.003]} \\
 & 20\% & 0.016 & 0.008 & 0.016 & 0.016 & 0.014 & 0.008 {\tiny[0.004, 0.012]} & $-$0.001 {\tiny[$-$0.004, 0.004]} & $-$0.000 {\tiny[$-$0.005, 0.007]} & 0.002 {\tiny[$-$0.003, 0.009]} \\
 & 30\% & 0.020 & 0.012 & 0.019 & 0.022 & 0.018 & 0.008 {\tiny[0.004, 0.013]} & 0.001 {\tiny[$-$0.003, 0.006]} & $-$0.002 {\tiny[$-$0.006, 0.002]} & 0.002 {\tiny[$-$0.003, 0.009]} \\
 & 50\% & 0.027 & 0.019 & 0.020 & 0.031 & 0.023 & 0.008 {\tiny[0.002, 0.015]} & 0.007 {\tiny[0.003, 0.011]} & $-$0.004 {\tiny[$-$0.009, 0.000]} & 0.004 {\tiny[$-$0.003, 0.013]} \\
\midrule
Audio & 10\% & 0.060 & 0.023 & 0.090 & 0.087 & 0.068 & 0.037 {\tiny[0.025, 0.050]} & $-$0.030 {\tiny[$-$0.038, $-$0.023]} & $-$0.027 {\tiny[$-$0.035, $-$0.019]} & $-$0.008 {\tiny[$-$0.013, $-$0.003]} \\
 & 20\% & 0.077 & 0.034 & 0.100 & 0.099 & 0.081 & 0.044 {\tiny[0.031, 0.057]} & $-$0.022 {\tiny[$-$0.031, $-$0.015]} & $-$0.022 {\tiny[$-$0.031, $-$0.015]} & $-$0.004 {\tiny[$-$0.009, 0.000]} \\
 & 30\% & 0.074 & 0.037 & 0.092 & 0.096 & 0.083 & 0.037 {\tiny[0.026, 0.050]} & $-$0.017 {\tiny[$-$0.023, $-$0.012]} & $-$0.021 {\tiny[$-$0.028, $-$0.015]} & $-$0.009 {\tiny[$-$0.013, $-$0.004]} \\
 & 50\% & 0.057 & 0.041 & 0.064 & 0.071 & 0.061 & 0.016 {\tiny[0.008, 0.025]} & $-$0.006 {\tiny[$-$0.011, $-$0.002]} & $-$0.014 {\tiny[$-$0.018, $-$0.010]} & $-$0.003 {\tiny[$-$0.007, 0.001]} \\
\midrule
Image--text & 1 quote & 0.285 & 0.142 & 0.317 & 0.311 & 0.298 & 0.143 {\tiny[0.115, 0.174]} & $-$0.032 {\tiny[$-$0.040, $-$0.025]} & $-$0.026 {\tiny[$-$0.034, $-$0.019]} & $-$0.013 {\tiny[$-$0.021, $-$0.006]} \\
 & 2 quotes & 0.351 & 0.216 & 0.368 & 0.389 & 0.374 & 0.135 {\tiny[0.107, 0.164]} & $-$0.018 {\tiny[$-$0.033, $-$0.001]} & $-$0.039 {\tiny[$-$0.053, $-$0.022]} & $-$0.023 {\tiny[$-$0.033, $-$0.012]} \\
 & 3 quotes & 0.379 & 0.257 & 0.380 & 0.431 & 0.405 & 0.122 {\tiny[0.098, 0.147]} & $-$0.002 {\tiny[$-$0.022, 0.024]} & $-$0.052 {\tiny[$-$0.075, $-$0.034]} & $-$0.027 {\tiny[$-$0.050, $-$0.008]} \\
\bottomrule
\end{tabular}
}
\caption{Lift of every method at every budget on Qwen2.5-Omni-7B, and the paired difference between OmniTrace and each baseline with its 95\% bootstrap interval. OT-trained: OmniTrace-trained (\Cref{appendix:trained}).}
\label{tab:budget_base}
\end{table*}

\Cref{tab:budget} gives the drops, lifts, and keep gains of OmniTrace at every budget, and \Cref{tab:budget_base} the comparison with the baselines.
The share of leave-one-out's lift that OmniTrace reaches grows with the budget in every modality.
On video at the 10\% budget, the keep gain against embedding similarity is dominated by one question, in which keeping the baseline's units costs 11 nats per token.

\subsection{Predicted Labels}
\label{appendix:labels}

\begin{table*}[t]
\centering
\small
\setlength{\tabcolsep}{4pt}
\resizebox{\textwidth}{!}{
\begin{tabular}{llcccccc}
\toprule
\textbf{Model} & \textbf{Modality} ($n$) & Units labelled & No label & Lift & Keep gain & S-lift & S-spec \\
\midrule
Qwen2.5-Omni-7B & Video (244) & 4.0 (10\%) & 2\% & 0.006 {\tiny[0.003, 0.010]} & 0.046 {\tiny[0.011, 0.111]} & 0.012 {\tiny[0.009, 0.014]} & 0.007 {\tiny[0.004, 0.010]} \\
 & Audio (196) & 2.6 (16\%) & 14\% & 0.079 {\tiny[0.060, 0.098]} & 0.077 {\tiny[0.060, 0.096]} & 0.125 {\tiny[0.107, 0.143]} & 0.106 {\tiny[0.082, 0.130]} \\
 & Image--text (400) & 3.1 (21\%) & 0\% & 0.364 {\tiny[0.329, 0.403]} & 0.393 {\tiny[0.353, 0.438]} & 0.411 {\tiny[0.387, 0.434]} & 0.240 {\tiny[0.214, 0.267]} \\
\midrule
MiniCPM-o 4.5 & Video (233) & 5.2 (13\%) & 0\% & 0.006 {\tiny[0.002, 0.010]} & 0.006 {\tiny[$-$0.004, 0.015]} & 0.014 {\tiny[0.008, 0.020]} & 0.008 {\tiny[0.001, 0.015]} \\
 & Audio (187) & 1.6 (15\%) & 0\% & 0.073 {\tiny[0.047, 0.100]} & 0.042 {\tiny[0.024, 0.063]} & 0.080 {\tiny[0.059, 0.100]} & 0.047 {\tiny[0.011, 0.085]} \\
 & Image--text (400) & 3.1 (21\%) & 0\% & 0.170 {\tiny[0.140, 0.201]} & 0.149 {\tiny[0.124, 0.178]} & 0.458 {\tiny[0.437, 0.479]} & 0.344 {\tiny[0.320, 0.369]} \\
\midrule
Qwen3-Omni-30B-A3B & Video (246) & 5.1 (12\%) & 2\% & 0.029 {\tiny[0.022, 0.037]} & 0.017 {\tiny[0.009, 0.024]} & 0.029 {\tiny[0.023, 0.035]} & 0.013 {\tiny[0.007, 0.020]} \\
 & Audio (196) & 2.7 (18\%) & 11\% & 0.125 {\tiny[0.099, 0.154]} & 0.112 {\tiny[0.088, 0.140]} & 0.105 {\tiny[0.091, 0.121]} & 0.079 {\tiny[0.058, 0.101]} \\
 & Image--text (400) & 3.3 (22\%) & 0\% & 0.368 {\tiny[0.346, 0.389]} & 0.380 {\tiny[0.357, 0.404]} & 0.480 {\tiny[0.466, 0.496]} & 0.279 {\tiny[0.261, 0.297]} \\
\bottomrule
\end{tabular}
}
\caption{Causal test of the labels of \Cref{eq:label} at $\tau=1.5$. Units labelled: mean number per response and share of the candidate units. No label: responses whose response-level label is empty. Lift and keep gain compare the labelled units with random units of the same count.}
\label{tab:labels}
\end{table*}

\Cref{tab:labels} tests the labelled sets.
All lifts, sentence lifts, and sentence specificities are above zero.
The keep gain on video is positive but not reliable for MiniCPM-o~4.5, and for Qwen2.5-Omni its mean is raised by one question.

\subsection{Label Threshold}
\label{appendix:threshold}

For image--text input, $\tau=1.5$ gives the highest mean F1 against the gold quotes on 200 development questions that are disjoint from the test questions.
Audio and video have no usable labels for this, so we compared thresholds by the causal effect of the labelled sets on a random fifth of the questions (54 video, 53 audio), which is excluded from all results.
No threshold was reliably better than 1.5 there; lower values scored slightly higher, so this check supports 1.5 as a safe choice and does not show that it is optimal.
MiniCPM-o~4.5 uses the same value without any check.

\subsection{Variants and Human-Evidence Units}
\label{appendix:variants}

\begin{table*}[t]
\centering
\small
\setlength{\tabcolsep}{4.2pt}
\resizebox{\textwidth}{!}{
\begin{tabular}{lcccccccccccc}
\toprule
 & \multicolumn{4}{c}{\textbf{Video}} & \multicolumn{4}{c}{\textbf{Audio}} & \multicolumn{4}{c}{\textbf{Image--text}} \\
\cmidrule(lr){2-5}\cmidrule(lr){6-9}\cmidrule(lr){10-13}
 & Lift & Keep & S-lift & S-spec & Lift & Keep & S-lift & S-spec & Lift & Keep & S-lift & S-spec \\
\midrule
\multicolumn{13}{>{\columncolor{gray!15}}l}{\textit{Qwen2.5-Omni-7B} \;($n$ = 246, 196, 400)}\\
\textbf{OmniTrace} & 0.016 & 0.010 & 0.015 & 0.007 & 0.069 & 0.066 & 0.103 & 0.084 & 0.348 & 0.412 & 0.375 & 0.245 \\
\;argmax tracing + curation & 0.009\mdn & 0.008 & 0.010\mdn & 0.004\mdn & 0.067 & 0.064 & 0.096\mdn & 0.081 & 0.346 & 0.420 & -- & -- \\
\;last layer only & 0.014 & 0.010 & -- & -- & 0.048\mdn & 0.046\mdn & -- & -- & 0.306\mdn & 0.380\mdn & -- & -- \\
\;per-token mean & -- & -- & -- & -- & -- & -- & -- & -- & 0.224\mdn & 0.255\mdn & 0.286\mdn & 0.214\mdn \\
Human-evidence units & 0.009\mdn & -- & -- & -- & -- & -- & -- & -- & 0.211\mdn & -- & -- & -- \\
\midrule
\multicolumn{13}{>{\columncolor{gray!15}}l}{\textit{MiniCPM-o 4.5} \;($n$ = 233, 187, 400)}\\
\textbf{OmniTrace} & 0.027 & 0.026 & 0.029 & 0.013 & 0.143 & 0.106 & 0.120 & 0.055 & 0.105 & 0.074 & 0.158 & 0.098 \\
\;argmax tracing + curation & 0.027 & $-$0.051\mdn & 0.025 & $-$0.001\mdn & 0.020\mdn & 0.001\mdn & 0.021\mdn & 0.006\mdn & 0.079\mdn & 0.058\mdn & -- & -- \\
\;last layer only & 0.025 & 0.005\mdn & -- & -- & 0.091\mdn & 0.055\mdn & -- & -- & 0.092 & 0.056 & -- & -- \\
\;per-token mean & -- & -- & -- & -- & -- & -- & -- & -- & 0.083 & 0.057 & 0.188\mup & 0.165\mup \\
Human-evidence units & 0.043\mup & -- & -- & -- & -- & -- & -- & -- & 0.334\mup & -- & -- & -- \\
\midrule
\multicolumn{13}{>{\columncolor{gray!15}}l}{\textit{Qwen3-Omni-30B-A3B} \;($n$ = 246, 196, 400)}\\
\textbf{OmniTrace} & 0.033 & 0.016 & 0.032 & 0.011 & 0.112 & 0.097 & 0.095 & 0.065 & 0.316 & 0.304 & 0.448 & 0.309 \\
\;argmax tracing + curation & 0.020\mdn & 0.022\mup & 0.019\mdn & 0.007\mdn & 0.112 & 0.105\mup & 0.087\mdn & 0.068 & 0.323 & 0.322\mup & -- & -- \\
\;last layer only & 0.008\mdn & 0.000\mdn & -- & -- & 0.045\mdn & 0.030\mdn & -- & -- & 0.125\mdn & 0.124\mdn & -- & -- \\
\;per-token mean & -- & -- & -- & -- & -- & -- & -- & -- & 0.240\mdn & 0.233\mdn & 0.381\mdn & 0.297\mdn \\
Human-evidence units & 0.027 & -- & -- & -- & -- & -- & -- & -- & 0.267\mdn & -- & -- & -- \\
\bottomrule
\end{tabular}
}
\caption{OmniTrace on three models, variants that change one design choice, and human-marked evidence.
Lift and Keep: lift and keep gain at response level (20\% of the duration; 2 quotes); S-lift and S-spec: sentence lift and specificity, in nats per token.
\mdn / \mup: reliably lower / higher than OmniTrace on the same questions or sentences.
\emph{Argmax tracing + curation} traces each token to its highest-attention position and filters the votes; \emph{last layer only} and \emph{per-token mean} change one choice in \Cref{eq:mass}.
\emph{Human-evidence units} hides the units annotators marked (not available for audio); their number differs from the budget, so their lift is against random sets of their own size.}
\label{tab:models}
\end{table*}

\begin{figure}[t]
\centering
\includegraphics[width=\colfigwidth]{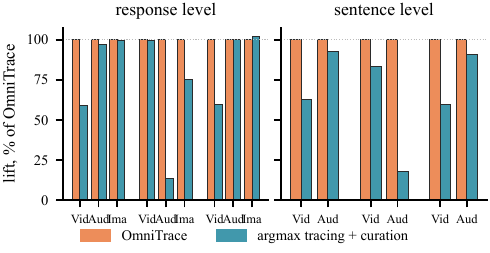}
\caption{Lift of argmax tracing with curation as a share of OmniTrace's lift. In each panel the groups of bars are, from left to right, Qwen2.5-Omni-7B, on which the variant was developed, MiniCPM-o~4.5, and Qwen3-Omni-30B.}
\label{fig:transfer}
\end{figure}

\Cref{tab:models} compares OmniTrace with variants that change one design choice, and \Cref{fig:transfer} shows the comparison with hard tracing.
On Qwen2.5-Omni, argmax tracing with curation is level with OmniTrace on audio and image--text input and behind on video.
On MiniCPM-o~4.5 it loses most of its effect on audio, is behind on image--text input, and on video its top units are worse to keep than random ones.
On Qwen3-Omni-30B it is level with OmniTrace in lift on audio and image--text input, slightly ahead in keep gain, and behind at sentence level on audio; on video it keeps 60\% of the lift, as on Qwen2.5-Omni.
Using the last layer alone is reliably worse on audio for all three models, and on Qwen3-Omni it loses more than half of the lift in every modality.

The last rows hide the units that annotators marked.
On Qwen2.5-Omni, the units OmniTrace selects have a larger lift than the human-marked ones.
On Qwen3-Omni-30B the same holds for image--text input (0.316 against 0.267), and on video the two are close (0.033 against 0.027).
On MiniCPM-o~4.5 the order is reversed, most clearly on image--text input: there OmniTrace finds units that matter, but not the ones that matter most.

\subsection{Unit Size}
\label{appendix:size}

\begin{table}[t]
\centering
\tablesize
\setlength{\tabcolsep}{3.5pt}
\begin{tabular}{lccc}
\toprule
\textbf{Ranking} & $\rho$(score, size) & Top-1 size & Image share\\
\midrule
OmniTrace (sum) & $+$0.51 & 1.94 & 0.59\\
Per-token mean & $-$0.60 & 0.44 & 0.11\\
Leave-one-out & $+$0.09 & 1.62 & 0.48\\
ContextCite & $+$0.14 & 1.65 & 0.52\\
OmniTrace-trained & $+$0.08 & 1.71 & 0.52\\
Human evidence & $+$0.25 & 1.85 & 0.67\\
\bottomrule
\end{tabular}
\caption{Unit size on image--text input (Qwen2.5-Omni-7B). $\rho$: rank correlation between a quote's score and its token count, per question. Top-1 size: tokens of the top quote over the mean quote size. Image share: share of images among the top-ranked quotes.}
\label{tab:size}
\end{table}

\Cref{tab:size} gives the numbers behind \Cref{sec:size}. The per-token mean ranks short passages first and rarely an image. On MiniCPM-o~4.5 the mean is better than the sum at sentence level (\Cref{tab:models}), so the best correction for size depends on the model.

\subsection{Unit Granularity}
\label{appendix:granularity}

The score of a merged unit is the sum of its parts, but a selected set can still change with the unit length.
We merge one-second units into two- and four-second units and select the top 20\% of the duration again.
On video the selected time overlaps the one-second selection with IoU 0.60 and 0.50, where random rankings give 0.15; on audio the values are 0.53 and 0.42.
The selection is far from arbitrary but not invariant, so the unit length should be fixed and reported.

\subsection{Attribution Strength}
\label{appendix:flagging}

At sentence level, the attention mass a sentence places on the sources correlates with how much the sentence depends on them (rank correlation 0.60 on audio and image--text input, 0.26 on video, Qwen2.5-Omni).
At response level, a low share of mass on the audio flags wrong answers on Qwen2.5-Omni (AUROC 0.74), but not in the other modalities and not on MiniCPM-o~4.5.
Attribution strength thus indicates whether a sentence draws on the sources; it is not a general detector of wrong answers.

\section{Answer Accuracy of Long Answers}
\label{appendix:judge}

MMDocRAG answers are free text, so we grade them with an LLM judge (\texttt{claude-opus-5-5}).
The judge sees the question, the short and long reference answers, and the model's response, and returns \emph{correct} (all facts the reference needs), \emph{partial} (some facts right, at least one missing or wrong), or \emph{incorrect}. It does not see the quotes or images.
For Qwen2.5-Omni-7B, 374 of the 400 responses received a verdict: 23.8\% correct, 43.0\% partial, 33.2\% incorrect.
For MiniCPM-o~4.5 (400 verdicts): 50.5\% correct and 35.3\% partial.
For Qwen3-Omni-30B (400 verdicts): 54.0\% correct and 32.8\% partial.
The accuracy is used only to describe the models and to select correctly answered questions in \Cref{sec:agreement}.

\section{Data and Code}
We will release the code of OmniTrace, the causal tests, and the baselines, together with the question lists and the model responses used in this paper. The datasets keep the licenses of their creators.

\end{document}